\documentclass{ecai2014}
\usepackage{times}
\usepackage{latexsym}
\usepackage{url}
\usepackage{xspace}
\usepackage{amsmath}
\usepackage{amssymb}
\usepackage{microtype}
\usepackage{tikz}

\newcommand{\mi}[1]{\ensuremath{\mathit{#1}}}

\newcommand{\define}[1]{\emph{#1}}

\newcommand{\la}{\leftarrow}

\newcommand{\setbs}[0]{\ensuremath{\mathcal{BS}}\xspace}
\newcommand{\setkb}[0]{\ensuremath{\mathcal{KB}}\xspace}
\newcommand{\setacc}[0]{\ensuremath{\mathcal{ACC}}\xspace}
\newcommand{\accN}[0]{\ensuremath{\mathrm{ACC}}\xspace}

\newcommand{\lkb}[1]{\ensuremath{\setkb_{#1}}\xspace}
\newcommand{\lbs}[1]{\ensuremath{\setbs_{#1}}\xspace}
\newcommand{\lacc}[1]{\ensuremath{\setacc_{#1}}\xspace}
\newcommand{\bs}[0]{\ensuremath{S}\xspace}

\newcommand{\relout}[4]{\ensuremath{{d_{#1}}({#2,#3,#4})}\xspace}

\newcommand{\contextLS}[1]{\ensuremath{\ls_{#1}}\xspace}
\newcommand{\contextName}[1]{\ensuremath{\cname_{#1}}\xspace}

\newcommand{\attime}[2]{\ensuremath{#1^{#2}}}

\newcommand{\contexts}[0]{\ensuremath{\mathsf{C}}\xspace}
\newcommand{\onames}[0]{\ensuremath{\mathsf{O}}\xspace}

\newcommand{\conf}[0]{\ensuremath{\mathit{cf}}\xspace}
\newcommand{\sysconf}[0]{\ensuremath{\mathit{Cf}}\xspace}

\newcommand{\busy}[0]{\ensuremath{\mathit{busy}}\xspace}
\newcommand{\eoc}[0]{\ensuremath{\mathrm{EOC}}\xspace}

\newcommand{\C}[1]{\ensuremath{C_{#1}}\xspace}

\newcommand{\IL}[0]{\ensuremath{\mathcal{IL}}\xspace}
\newcommand{\ls}[0]{\ensuremath{\mathcal{LS}}\xspace}
\newcommand{\outr}[0]{\ensuremath{\mathrm{OR}}\xspace}
\newcommand{\ib}[0]{\ensuremath{\mathrm{ib}}\xspace}
\newcommand{\ob}[0]{\ensuremath{\mathrm{ob}}\xspace}
\newcommand{\buf}[0]{\ensuremath{\mathrm{b}}\xspace}
\newcommand{\cuf}[0]{\ensuremath{\mathrm{cu}}\xspace}
\newcommand{\cc}[0]{\ensuremath{\mathrm{cc}}\xspace}
\newcommand{\cname}[0]{\ensuremath{\name}\xspace}

\newcommand{\names}[0]{\ensuremath{\mathcal{N}}\xspace}
\newcommand{\namesOfAMCS}[1]{\ensuremath{\names({#1})}\xspace}
\newcommand{\data}[0]{\ensuremath{\mathit{d}}\xspace}
\newcommand{\info}[0]{\ensuremath{\mathsf{i}}\xspace}
\newcommand{\outName}[0]{\ensuremath{\mathsf{o}}\xspace}
\newcommand{\name}[0]{\ensuremath{\mathsf{n}}\xspace}
\newcommand{\cm}[0]{\ensuremath{\mathit{cm}}\xspace}
\newcommand{\head}[1]{\ensuremath{\mi{hd}({#1})}\xspace}
\newcommand{\body}[1]{\ensuremath{\mathrm{bd}({#1})}\xspace}

\newcommand{\kb}[0]{\ensuremath{\mathrm{KB}}\xspace}

\newcommand{\naf}[0]{\ensuremath{\mathrm{not}}\xspace}

\newcommand{\tuple}[1]{\ensuremath{\tupleLeft {#1} \tupleRight}\xspace}
\newcommand{\tupleLeft}[0]{\ensuremath{\langle}\xspace}
\newcommand{\tupleRight}[0]{\ensuremath{\rangle}\xspace}

\newcommand{\integers}[0]{\ensuremath{{\mathbb{Z}}}\xspace}

\newcommand{\etal}[0]{\emph{et al.}\xspace}

\newcommand{\MCSinternal}{MCS}

\newcommand{\mcss}{\MCSinternal{}s\xspace}
\newcommand{\eMCSinternal}{eMCS}

\newcommand{\emcss}{\eMCSinternal{}s\xspace}
\newcommand{\aMCSinternal}{aMCS}
\newcommand{\amcs}{\aMCSinternal\xspace}
\newcommand{\amcss}{\aMCSinternal{}s\xspace}
\newcommand{\mmcs}{\mMCSinternal\xspace}
\newcommand{\mmcss}{\mMCSinternal{}s\xspace}
\newcommand{\rMCSinternal}{rMCS}
\newcommand{\rmcs}{\rMCSinternal\xspace}
\newcommand{\rmcss}{\rMCSinternal{}s\xspace}
\newcommand{\mMCSinternal}{mMCS}
\newcommand{\iec}[0]{i.e.,\xspace}
\newcommand{\cf}[0]{cf.\xspace}
\newcommand{\egc}[0]{e.g.,\xspace}

\newcommand{\bi}[0]{\begin{itemize}}
\newcommand{\ei}[0]{\end{itemize}}
\newcommand{\ttitem}[2]{\item [\texttt{#1}] #2}

\newtheorem{definition}{Definition}

\begin{document}

\title{Asynchronous Multi-Context Systems\thanks{This work has been partially supported by the German Research Foundation (DFG) under grants BR-1817/7-1 and FOR 1513.}}

\author{Stefan Ellmauthaler \and J\"org P\"uhrer\institute{Institute of Computer Science, Leipzig University,
Germany, email: \{ellmauthaler,puehrer\}@informatik.uni-leipzig.de} }

\maketitle
\bibliographystyle{ecai2014}

\begin{abstract}
In this work, we present \define{asynchronous multi-context systems} (\amcss), which provide a framework for loosely coupling different knowledge representation formalisms that allows for online reasoning in a dynamic environment.
Systems of this kind may interact with the outside world via input and output streams and may therefore react to a continuous flow of external information.
In contrast to recent proposals, contexts in an \amcs communicate with each other in an asynchronous way which fits the needs of many application domains and is beneficial for scalability.
The federal semantics of \amcss renders our framework an integration approach rather than a knowledge representation formalism itself.
We illustrate the introduced concepts by means of an example scenario dealing with rescue services.
In addition, we compare \amcss to reactive multi-context systems and describe how to simulate the latter with our novel approach.
%
% We present a new framework to model loosely connected knowledge representation frameworks, which communicate in an asynchronous way and may react to (possibly external) information on demand. The introduced concepts will be presented as a formal framework together with an example scenario.
% In addition, we will compare this framework to another recently introduced multi-context system framework and describe how to model it with our genuine approach.
\end{abstract}

\section{Introduction}
Research in the field of knowledge representation (KR) has originated a plethora of different languages and formats.
Based on these formal concepts a wealth of tools has emerged (\egc ontologies, triple-stores, modal logics, temporal logics, nonmonotonic logics, logic programs under nonmonotonic answer set semantics, \ldots).
In a \emph{``connected world''} it is desirable not to spread out information over different applications but to have it available for every application if need be.
Expressing all the knowledge usually represented in specifically tailored languages in a universal language would be too hard to achieve from the point of view of complexity as well as the troubles arising from the translation of the representations.
Instead, a framework seems desirable that integrates multiple existing formalisms in order to represent every piece of knowledge in the language that is most appropriate for it.

Another aspect that has received little attention in the development of many KR formalisms is that in a variety of applications, knowledge is provided in a constant flow of information and it is desired to reason over this knowledge in a continuous manner.
Many formalisms are conceptually one-shot formalisms: given a knowledge base, the user triggers the computation of a result (\egc the answer to a query).
In this paper we aim at using KR formalisms in an \emph{online} fashion as it has been done in recent works, \egc on stream data processing and querying~\cite{Zaniolo2012,Le-Phuoc2012}, stream reasoning with answer set programming~\cite{Gebser2012}, and forgetting~\cite{Lang2010,Cheng2006}.

To address the demand for an integration of heterogeneous knowledge representation formalisms together with the awareness of a continuous flow of knowledge over time, reactive multi-context systems (\rmcss)~\cite{Brewka2014} and evolving multi-context systems (\emcss)~\cite{GKL2014} where proposed.
Both frameworks are based on the ideas of managed multi-context systems (\mmcss)~\cite{BrewkaEFW11} which combine multiple \define{contexts} which can be seen as representations of different formalisms.
The semantics of \rmcss and \emcss are based on the notion of an \define{equilibrium} which realises a tight semantical integration of the different context formalisms which is in many applications not necessary.
Due to reasoning over all contexts, the whole computation is necessarily synchronous as the different contexts have to agree on common beliefs for establishing equilibria.

Many real world applications which utilise communication between different services use asynchronous communication protocols (\egc web services) and compute as soon as they have appropriate information about the problem they have to address.
Therefore, we introduce \emph{asynchronous multi-context systems} (\amcss), a framework for loosely coupled knowledge representation formalisms and services.
It still provides the capabilities to express different knowledge representation languages and the translation of information from one formalism to another.
In addition, \amcss are also aware of continuous streams of information and provide ways to model the asynchronous exchange of information.
To communicate with the environment, they utilise input and output streams.

% Together with a technical and formal approach to the topic, we will also propose our framework as a tool to model and describe already existing systems and problems.
% It shall help to identify where which kind of information shall be exchanged and visualise how the different parts of a specific instance of the framework will work with each other.

We will illustrate \amcss using the example of a task planner for medical rescue units.
Here, we assume a scenario where persons are calling an emergency response team to report incidents and the employee needs to collect all relevant information about the case.
Afterwards, the case needs to be classified and available resources (\egc free ambulances, \ldots) have to be assigned to the emergencies.
In addition, current traffic data as well as the estimated time of arrival should be considered by another employee, the dispatcher.
Our proposed \amcs that we understand as a recommender-system for the emergency response employee as well as to the dispatcher, incorporates different contexts like a medical ontology, a database with the current state of the ambulances, or a navigation system which is connected to a traffic density reporter.
We want to stress that this might be one application where it would be a great gain for the overall system by allowing asynchronous computation and communication such that it is not necessary to wait for all other contexts (\egc it would be unnecessary to wait for the recommendation of a plan of action for the dispatcher during the treatment of an emergency call). 

The remainder of this paper is structured as follows.
At first we will give a short background on concepts we need.
In Section~\ref{sec:amcss}, we extend the basic ideas of MCS to propose our new notion of \amcs for modelling asynchronous interaction between coupled knowledge representation formalisms and formally characterise its behaviour over time.
The subsequent section presents an example scenario, where asynchronous computation and a reactive response to different events is needed.
Section~\ref{sec:mmcs} compares \amcss to \rmcss and shows how the latter can be simulated by the former.
Section~\ref{sec:conclusion} concludes this paper with a discussion  including an outlook on future work.

\section{Preliminaries}\label{sec:prelim}
We base our approach on the underlying ideas of \mmcss~\cite{BrewkaEFW11} which
extend heterogeneous multi-context systems (\mcss)~\cite{BrewkaEF11} by a management layer.
It allows for complex updates and revisions of knowledge bases and is realised by
a management function that provides the updates for each equilibrium.
Despite we build on \mmcss, they differ substantially in some aspects from the formalism we introduce in this work for reasons intrinsic to the asynchronous approach (\cf~Section~\ref{sec:mmcs}).
Consequently, we only reuse basic notions from the original work and refer the interested reader to the paper of Brewka \etal~\cite{BrewkaEFW11} for full details on \mmcs.

Like \mmcs, \amcss build on an abstract notion of a \emph{logic suite} which can be seen as an abstraction of different formalisms for knowledge representation.
A logic suite is a triple $\ls=\tuple{\setkb,\setbs, \setacc}$, where $\setkb$ is the set of admissible knowledge bases (KBs) of $\ls$. 
Each knowledge base is a set of formulas that we do not further specify.
\setbs is the set of possible belief sets of \ls, whose elements are beliefs. 
A semantics for $\ls$ is a function $\accN : \mathit{KB} \to 2^{\setbs}$ assigning to each KB a set of acceptable belief sets.
Using a semantics with potentially more than one acceptable belief set allows for modelling non-determinism, where each belief set corresponds to an alternative solution.
Finally, $\setacc$ is a set of semantics for $\ls$.
We denote $\setkb$,$\setbs$, respectively, $\setacc$ by $\lkb{\ls}$,$\lbs{\ls}$, respectively, $\lacc{\ls}$.

The motivation behind having multiple semantics for one formalism is that in our framework, the semantics of a formalism can be changed over time.
While it is probably rarely the case that one wants to switch between different families of semantics during a run, \egc  from the stable-model semantics to the well-founded semantics of logic programs  other switches of semantics are quite natural to many applications:
we use different semantics to express different reasoning modes or to express different queries, \iec
$\accN_1$ returns belief sets answering a query $q_1$, whereas
$\accN_2$ answers query $q_2$; 
$\accN_3$, in turn, could represent the computation of all solutions to a problem, whereas at some point in time one could be interested in using $\accN_4$ that only computes a single solution.
For instance one that is optimal with respect to some criterion.

\section{Asynchronous Multi-Context Systems}\label{sec:amcss}
An \amcs is built up by multiple contexts  which are defined next and which are used for representing reasoning units.
We assume a set $\names$ of names that will serve as labels for sensors, contexts, and output streams.
\begin{definition}
A context is a pair $C=\tuple{\cname,\ls}$ where $\cname\in\names$ is the name of the context and \ls is a logic suite.
\end{definition}
\noindent
For a given context $C=\tuple{\cname,\ls}$ we denote \cname and \ls by $\contextName{C}$ and $\contextLS{C}$, respectively.

\begin{definition}
An \amcs (of length $n$ with $m$ output streams) is a pair $M=\tuple{\contexts,\onames}$, where 
$\contexts=\tuple{C_1,\dots,C_n}$ is an $n$-tuple of contexts
and $\onames=\tuple{\outName_1,\dots,\outName_m}$ with $\outName_j\in\names$ for each $1\leq j\leq m$ is a tuple containing the names of the output streams of $M$.
\end{definition}
By $\namesOfAMCS{M}$ we denote the set 
$\{\contextName{C_1},\dots,\contextName{C_n},\outName_1,\dots,\outName_m\}$ of names of contexts and output streams of $M$.

A context in an \amcs communicates with other contexts and the outside world by means of streams of data.
In particular, we assume that every context has an input stream on which information can be written from both external
sources (we call them sensors) and internal sources (\iec other contexts).
For the data in the communication streams we assume a communication language \IL where every $\info\in\IL$ is an abstract \define{piece of information}.
In our framework, the data in the input stream of a context and the data in output streams are modelled by information buffers that are defined in the following.
\begin{definition}
A \define{data package} is a pair $\data=\tuple{s,I}$, where $s\in\names$ is either a context name or a sensor name, stating the 
\define{source} of \data, and $I\subseteq\IL$ is a set of pieces of information.
An \define{information buffer} is a sequence of data packages.
\end{definition}

As we assume that data is asynchronously passed to a context on its input stream, it is natural that not all information required for a computation is available
at all times.
Consequently, we need means to decide whether a computation should take place, depending on the current KB and the data currently available on the stream,
or whether the context has to wait for more data.
In our framework, this decision is made by a computation controller as defined next.
\begin{definition}
Let $C= \tuple{\cname,\ls}$ be a context.
A \define{computation controller} for $C$ is a relation
\cc between a KB $\kb\in\lkb{\ls}$ and a finite information buffer.
\end{definition}
\noindent
Thus, if $\tuple{\kb,\ib}\in\cc$ then a computation should take place,
whereas $\tuple{\kb,\ib}\not\in\cc$ means that further information is required
before the next computation is triggered in the respective context.

In contrast to the original definition of multi-context systems~\cite{BrewkaE07} and extensions thereof,
we do not make use of so-called \define{bridge rules} as a means to communicate:
a bridge rule defines which information a context should obtain based on the results
of all the contexts of a multi-context system.
In the asynchronous approach, we do not have (synchronised) results of all contexts available in general.
As a consequence we use another type of rules, called \define{output rules}, that define which information should
be sent to another context or an output stream, based on a result of a single context.
\begin{definition}\label{def:or}
Let $C= \tuple{\cname,\ls}$ be a context.
An \define{output rule} $r$ for $C$ is an expression of the form \begin{align}
  \label{outputrule}
    \tuple{\name,\info} \la& b_1,\ldots, b_j,\naf\ b_{j+1},\ldots,\naf\ b_m,
\end{align}
such that $\name\in\names$ is the name of a context or an output stream, $\info\in\IL$ is a piece of information,
and every $b_\ell$ ($1 \leq \ell \leq m$) is a belief for $C$, \iec $b_\ell\in \bs$ for some $\bs\in\lbs{\ls}$.
\end{definition}
\noindent
We call $\name$ the \define{stakeholder} of $r$, \tuple{\name,\info} the head of $r$ denoted by \head{r} and $b_1,\ldots, b_j,\naf\ b_{j+1},\ldots,\naf\ b_m$ the body \body{r} of $r$.
Moreover, we say that $r$ is active under \bs, denoted by $\bs\models\body{r}$, if 
$\{b_1,\ldots, b_j\}\subseteq\bs$ and $\{b_{j+1},\ldots, b_m\}\cap\bs=\emptyset$.

Intuitively, the stakeholder is a reference to the addressee of information \info.
\begin{definition}
Let $C=\tuple{\cname,\ls}$ be a context, $\outr$ a set of output rules for $C$,
$\bs\in\lbs{\ls}$ a belief set, and  $\name'\in\names$ a name.
Then, the data package
$$
\relout{C}{\bs}{\outr}{\name'}=\tuple{\cname,\{i\mid r\in\outr, \head{r}=\tuple{\name',\info}, \bs\models\body{r}\}}
$$
is the \define{output} of $C$ with respect to \outr under \bs relevant for $\name$.
\end{definition}

Compared to previous notions of multi-context systems, 
contexts in our setting only specify which formalisms they use but
they do not contain knowledge bases, the concrete semantics to use, and communication specifications.
The reason is that for \amcss these may change over time.
Instead, we wrap concepts that are subject to change during runtime in the following notion of a
\define{configuration}.
\begin{definition}\label{def:config_context}
Let $C= \tuple{\cname,\ls}$ be a context.
A \define{configuration} of $C$ is a tuple $\conf=\tuple{\kb,\accN,\ib,\cm}$,
where 
$\kb\in\lkb{\ls}$,
$\accN\in\lacc{\ls}$,
\ib is a finite information buffer, and
\cm is a \define{context management} for $C$ which is a triple $\cm=\tuple{\cc,\cuf,\outr}$,
where
\bi
\item \cc is a computation controller for $C$,
\item \outr is a set of output rules for $C$, and
\item \cuf is a \define{context update function} for $C$ which is a function that maps an information buffer
$\ib=\data_1,\dots,\data_m$ and an admissible knowledge base of \ls
to a configuration $\conf'=\tuple{\kb',\accN',\ib',\cm'}$ of $C$
with $\ib'=\data_k,\dots,\data_m$ for some $k\ge 1$.
\ei
\end{definition}
\noindent 
We write $\cc_\cm$, $\cuf_\cm$, and $\outr_\cm$
to refer to the components of a given context management 
$\cm=\tuple{\cc,\cuf,\outr}$.
The context management is the counterpart of a \define{management function} of an \rmcs,
that computes an update of the knowledge base of a context given the results of bridge rules of the context.

In Section~\ref{sec:prelim} we already discussed why we want to change semantics over time.
Allowing also for changes of output rules can be motivated with applications where 
it should be dynamically decided where to direct the output of a context. 
For example, if a particular subproblem can be solved by two contexts $C_1$ and $C_2$
and it is known that some class of instances can be better solved by $C_1$ and others by $C_2$.
Then a third context that provides an instance can choose whether $C_1$ or $C_2$ should carry out the computation
by adapting its output rules.
Dynamically changing output rules and semantics could require adjustments of the other components
of the context management. Thus, it makes sense that also compution controllers and context update functions
are subject to change for the sake of flexibility.

\begin{definition}\label{def:config_amcs}
Let $M=\tuple{\tuple{C_1,\dots,C_n},\tuple{\outName_1,\dots,\outName_m}}$ be an \amcs.
A \define{configuration} of $M$ is a pair $$\sysconf=\tuple{\tuple{\conf_1,\dots,\conf_n},\tuple{\ob_1,\dots,\ob_m}},$$
where 
\bi 
 \item for all $1 \leq i \leq n$ $\conf_i=\tuple{\kb,\accN,\ib,\cm}$ is a configuration for $C_i$
   and for every output rule $r\in\outr_\cm$ we have $\name\in\namesOfAMCS{M}$ for $\tuple{\name,\info}=\head{r}$, and
 \item $\ob_j=\dots,\data_{l-1},\data_l$ is an information buffer with a final element $\data_l$ that corresponds to the data on the output stream named $\outName_j$ for each $1 \leq j \leq m$ such that for each $h\leq l$ with $\data_{h}=\tuple{\name,\info}$ we have $\name=\contextName{C_i}$ for some $1\leq i \leq n$.
\ei
\end{definition}

\begin{figure}[!thb]
  \centering
  \resizebox{\columnwidth}{!}{\begin{tikzpicture}
  \tikzstyle{ttext}=[font=\tiny,anchor=base]
  % amcs
  \draw (1,2) rectangle (8,7.5);
  \draw (6,7.1) rectangle  node {\amcs $M$} (8,7.5);
  % sensors
  \draw [fill=black] (0,7) circle (0.05);
  \draw (0,7) circle (0.2) node[label={[label distance=1]$s_1$}] (s1) {};
  \draw [fill=black] (0,6) circle (0.05);
  \draw (0,6) circle (0.2) node[label={[label distance=1]$s_2$}] (s2) {};
  \draw [fill=black] (0,5) circle (0.05);
  \draw (0,5) circle (0.2) node[label={[label distance=1]$s_3$}] (s3) {};
  % output
  \draw (9,5) rectangle (9.25,5.25) node [yshift=-3.5] (out1) {};
  \draw (9.25,5.25) -- (9.75,5.25);
  \draw (9,5) -- (9.5,5);

  \draw (9,4) rectangle (9.25,4.25) node [yshift=-3.5] (out2) {};
  \draw (9.25,4.25) -- (9.75,4.25);
  \draw (9,4) -- (9.5,4);

  \draw (9,3) rectangle (9.25,3.25) node [yshift=-3.5] (out3) {};
  \draw (9.25,3.25) -- (9.75,3.25);
  \draw (9,3) -- (9.5,3);
  % context
  \draw [rounded corners] (5.1,2.5) rectangle (6.6,4.1);
  \draw [fill=white,rounded corners] (5,3) rectangle node{$\C{3}$} (6,4);
  \node [ttext](kb3) at (6.29,3.6) {$\kb$};
  \node [ttext](acc3) at (6.29,3.2) {$\accN$};
  \node [ttext](ib3) at (5.22,2.7) {$\ib$};
  \node [ttext](cc3) at (5.55,2.7) {$\cc$};
  \node [ttext](cuf3) at (5.88,2.7) {$\cuf$};
  \node [ttext](or3)  at (6.29,2.7) {$\outr$};

  \draw [rounded corners] (2.1,5.5) rectangle (3.6,7.1);
  \draw [fill=white,rounded corners] (2,6) rectangle node{$\C{1}$} (3,7);
  \node [ttext](kb1) at (3.29,6.6) {$\kb$};
  \node [ttext](acc1) at (3.29,6.2) {$\accN$};
  \node [ttext](ib1) at (2.22,5.7) {$\ib$};
  \node [ttext](cc1) at (2.55,5.7) {$\cc$};
  \node [ttext](cuf1) at (2.88,5.7) {$\cuf$};
  \node [ttext](or1)  at (3.29,5.7) {$\outr$};

  \draw [rounded corners] (4.1,4.5) rectangle (5.6,6.1);
  \draw [fill=white,rounded corners] (4,5) rectangle node{$\C{2}$} (5,6);
  \node [ttext](kb2) at (5.29,5.6) {$\kb$};
  \node [ttext](acc2) at (5.29,5.2) {$\accN$};
  \node [ttext](ib2) at (4.22,4.7) {$\ib$};
  \node [ttext](cc2) at (4.55,4.7) {$\cc$};
  \node [ttext](cuf2) at (4.88,4.7) {$\cuf$};
  \node [ttext](or2)  at (5.29,4.7) {$\outr$};
  % communication flow
  \draw [->,dashed] (s1) -- (ib1);
  \draw [->,dashed] (s2) -- (ib1);
  \path [->,bend right,dashed] (s3) edge (ib3);
  \path [->,bend right] (or1) edge (ib2);
  \path [->, out=230, in=160] (or2) edge (ib3);
  \draw [->,shorten <= -0.06cm] (or3) -- (out3);
  \draw [->,shorten <= -0.06cm] (or3) -- (out2);
  \draw [->,shorten <= -0.06cm] (or2) -- (out1);
\end{tikzpicture}}
  \caption{An \amcs with three contexts, three sensors on the left side, and three output streams on the right side. 
          A solid line represents a flow of information from a context to its stakeholder streams, whereas a dashed line 
          indicates sensor data written to the input buffer of a context.}\label{fig:amcs}
\end{figure}
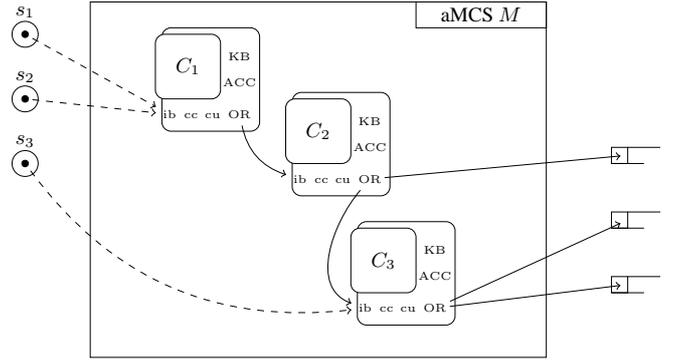

Figure~\ref{fig:amcs} depicts an \amcs $M$ with three contexts and a configuration for $M$.

We next characterise the dynamic behaviour of an \amcs.
For easier notation we stick to a discrete notion of time represented by integers.

\begin{definition}
Let $M=\tuple{\tuple{C_1,\dots,C_n},\tuple{\outName_1,\dots,\outName_m}}$ be an \amcs.
A run structure for $M$ is a sequence 
$$R=\dots,\attime{\sysconf}{t},\attime{\sysconf}{t+1},\attime{\sysconf}{t+2},\dots\quad\mbox{,}$$
where $t\in\integers$ is a point in time, and every $\attime{\sysconf}{t'}$
in $R$ ($t'\in\integers$) is a configuration of $M$.
\end{definition}
\noindent
We will sometimes use $\attime{\conf_i}{t}$ to denote the configuration of a context $i$ that appears at time $t$ in 
a given run structure in the context of a given \amcs.
Similarly, $\attime{\ob_j}{t}$ refers to the information buffer representing the data in the output stream named $\outName_j$.
Moreover, we write $\attime{\kb_i}{t}$, $\attime{\accN_i}{t}$, $\attime{\ib_i}{t}$, and $\attime{\cm_i}{t}$
to refer to the components of $\attime{\conf_i}{t}=\tuple{\kb,\accN,\ib,\cm}$.
We say that context $C_i$ is \define{waiting} at time $t$
if $\tuple{\attime{\kb_i}{t},\attime{\ib_i}{t}}\not\in\cc_{\attime{\cm_i}{t}}$.

\paragraph{From run structure to run}
In \amcss we take into account that the computation of the semantics of a knowledge base
needs time.
Moreover, in a computation of our framework, different belief sets
may become available at different times and verifying the non-existence of further belief sets can also take time
after the final belief set has been computated.
In order to model whether a context is busy with computing, 
we introduce a boolean variable $\attime{\busy_i}{t}$ 
for each configuration $\attime{\conf_i}{t}$ in a run structure.
Hence,
context $C_i$ is \define{busy} at time $t$ iff $\attime{\busy_i}{t}$ is true.
While a context is busy, it does not read new information from its input stream
until every belief set has been computed and it has concluded that no further belief set exists.

After the computation of a belief set, the output rules are applied in order to determine
which data is passed on to stakeholder contexts or output streams.
These are represented by \define{stakeholder buffers}:
An information buffer \buf is the stakeholder buffer of $C_i$ (for $\name$) at time $t$ if
\bi
\item $\buf=\attime{\ib_{i'}}{t}$ for some $1\leq i'\leq n$ such that $\name=\contextName{C_i}$ is stakeholder of some output rule in $\outr_{\attime{\cm_i}{t}}$ or
\item $\buf=\attime{\ob_{j'}}{t}$ for some $1\leq j'\leq m$ such that $\name=\outName_{j'}$ is stakeholder of some output rule in $\outr_{\attime{\cm_i}{t}}$.
\ei

In order to indicate that a computation has finished we assume a dedicated
symbol $\eoc\in\IL$ that notifies a context's stakeholder buffers about the end of a computation.

Next, we formally characterise the behaviour of \amcss followed by a summary of its intuition.
\begin{definition}\label{def:run}
Let $M$ be an \amcs of length $n$ with $m$ output streams and $R$ a run structure for $M$.
$R$ is a \define{run} for $M$ if the following conditions hold for every $1\leq i\leq n$ and every $1\leq j\leq m$:
\bi
\item[(i)] if $\attime{\conf_i}{t}$ and $\attime{\conf_i}{t+1}$ are defined, $C_i$ is neither busy nor waiting at time $t$,
      then 
  \bi
  \item $C_i$ is busy at time $t+1$,
  \item $\attime{\conf_i}{t+1}=\cuf_{\attime{\cm_i}{t}}(\attime{\ib_i}{t},\attime{\kb_i}{t})$
  \ei
  We say that $C_i$ \define{started a computation} for \attime{\kb_i}{t+1} at time $t+1$.
\item[(ii)] if $C_i$ \define{started a computation} for \kb at time $t$ then
\bi
  \item we say that this computation ended at time $t'$, if $t'$ is the earliest time point with $t'\ge t$ such that
        $\tuple{\contextName{C_i},\eoc}$ is added to every stakeholder buffer \buf of $C_i$ at $t'$;
        the addition of $\relout{C_i}{\bs}{\outr_{\attime{\cm_i}{t''}}}{\name}$ to \buf is called an \define{end of computation notification}.
  \item for all $t'>t$ such that $\attime{\conf_i}{t'}$ is defined, $C_i$ is busy at $t'$ unless the computation
        ended at some time $t''$ with $t<t''<t'$.
  \item if the computation ended at time $t'$ and $\attime{\conf_i}{t'+1}$ is defined then $C_i$ is not busy at $t'+1$.
\ei  
\item[(iii)] if $C_i$ \define{started a computation} for \kb at time $t$ that ended at time $t'$ then
      for every belief set $\bs\in\attime{\accN_i}{t}$ there is some time $t''$ with $t\leq t''\leq t'$ such that
  \bi
    \item $\relout{C_i}{\bs}{\outr_{\attime{\cm_i}{t''}}}{\name}$ is added to every stakeholder buffer \buf of $C_i$ for $\name$ at $t''$.
  \ei  
We say that $C_i$ \define{computed} \bs at time $t''$.
The addition of $\relout{C_i}{\bs}{\outr_{\attime{\cm_i}{t''}}}{\name}$ to \buf is called a \define{belief set notification}.

\item[(iv)] if $\attime{\ob_j}{t}$ and $\attime{\ob_j}{t+1}$ are defined
      and $\attime{\ob_j}{t}=\dots,\data_{l-1},\data_l$
      then $\attime{\ob_j}{t+1}=\dots,\data_{l-1},\data_l,\dots,\data_{l'}$ for some $l'\ge l$.
      Moreover, every data package $\data_{l''}$ with $l< l''\leq l'$ that was added at time $t+1$ results from 
      an end of computation notification or a belief set notification.
\item[(v)] if $\attime{\conf_i}{t}$ and $\attime{\conf_i}{t+1}$ are defined, $C_i$ is busy or waiting at time $t$,
      and $\attime{\ib_i}{t}=\data_1,\dots,\data_l$
      then we have $\attime{\ib_i}{t+1}=\data_1,\dots,\data_l,\dots,\data_{l'}$ for some $l'\ge l$.
      Moreover, every data package $\data_{l''}$ with $l< l''\leq l'$ that was added at time $t+1$ 
      either results from an end of computation notification or a belief set notification or 
      $\name\notin\namesOfAMCS{M}$ (\iec $\name$ is a sensor name) for $\data_{l''}=\tuple{\name,\info}$.
\ei
\end{definition}
Condition~(i) describes the transition from an idle phase to an ongoing computation.
The end of such a compation is marked by an end of computation notification
as introduced in Item~(ii).
Condition~(iii) states that between the start and the end of a computation
all belief sets are computed and stakeholders are notified.
Items~(iv) and (v) express how data is added to an output stream or to an input stream, respectively.
Note that here, sensors and the flow of information from sensors to the input buffers of contexts
are implicit.
That is, data packages from a sensor may appear at the end of input buffers at all times
and the only reference to a particular sensor is its name appearing in a data package.

Summarising the behaviour characterised by a run,
whenever a context $C$ is not busy, its context controller \cc 
checks whether a new computation should take place,
based on the knowledge base and the current input buffer of $C$.
If yes, the current configuration of the context is replaced by
a new one, computed by the context update function \cuf of $C$.
Here, the new input buffer has to be a suffix of the old one
and a new computation for the updated knowledge base starts. 
After an undefined period of time,
belief sets are computed and based on the application of output rules of $C$,
data packages are sent to stakeholder buffers.
At some point in time, when all belief sets have been computed,
an end of computation notification is sent to stakeholders, and the context
is not busy anymore.

\section{Scenario: Computer-Aided Emergency Team Management}

Now we want to consider a scenario, where \amcss may be used to describe the asynchronous information-exchange between different specialised reasoning systems.
Our example deals with a recommender-system for the coordination and handling of ambulance assignments.
The suggested \amcs supports decisions in various stages of an emergency case.
It gives assistance during the rescue call, helps in assigning priorities and rescue units to a case, and 
assists in the necessary communication among all involved parties.
The suggestions given by the system are based on different specialised systems which react to sensor readings. 
Moreover, the system can tolerate and incorporate overriding solutions proposed by the user that it considers non-optimal.
 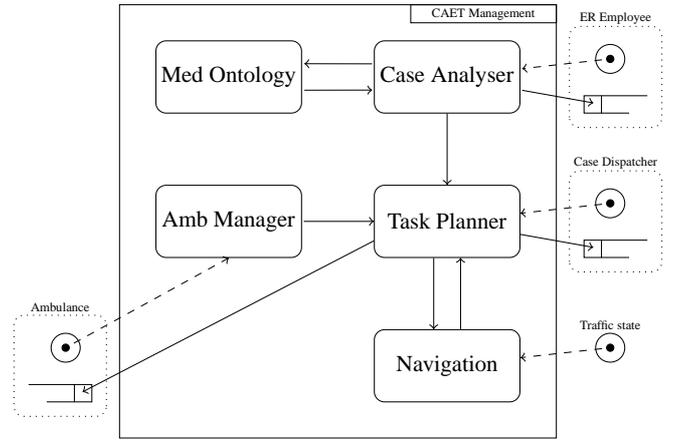
\begin{figure}[!htb]
   \centering
   \resizebox{\columnwidth}{!}{\begin{tikzpicture}
  \tikzstyle{context}=[rounded corners]
  \tikzstyle{senI}=[fill=black]
  \tikzstyle{bnode}=[rectangle,minimum width=2cm,minimum height=1cm]
  \tikzstyle{sonode}=[rectangle,minimum width=1.07cm, minimum height=1.4cm, font=\tiny,above,yshift=0.11cm]

  % context
  \draw [context] (3,0) rectangle node (nav) [bnode] {Navigation} (5,1);
  \draw [context] (0,2) rectangle node (rum) [bnode]{Amb Manager} (2,3);
  \draw [context] (3,2) rectangle node (tp) [bnode] {Task Planner} (5,3);
  \draw [context] (3,4) rectangle node (ca) [bnode] {Case Analyser} (5,5);
  \draw [context] (0,4) rectangle node (mdb) [bnode]{Med Ontology} (2,5);

  % sensors & output
  \draw [senI] (6.24,4.75) circle (0.05);
  \draw (6.24,4.75) circle (0.2) node (seme){};
  
  \draw (5.88,4) rectangle (6.12,4.24) node [yshift=-0.12cm] (oeme) {};
  \draw (6.12,4.24) -- (6.75,4.24);
  \draw (6.12,4) -- (6.5,4);

  \draw [context, dotted] (5.68,3.8) rectangle node [sonode] {ER Employee}(6.95,5.2);
  %--
  \draw [senI] (6.24,2.75) circle (0.05) node (scd) {};
  \draw (6.24,2.75) circle (0.2);
  
  \draw (5.88,2) rectangle (6.12,2.24) node [yshift=-0.12cm] (ocd) {};
  \draw (6.12,2.24) -- (6.75,2.24);
  \draw (6.12,2) -- (6.5,2);

  \draw [context, dotted] (5.68,1.8) rectangle node [sonode] {Case Dispatcher}(6.95,3.2);
  %--
  \draw [senI] (6.24,0.75) circle (0.05);
  \draw (6.24,0.75) circle (0.2) node[label={[label distance=1,font=\tiny]Traffic state}] (sts) {};
  %--
  \draw [senI] (-1.24,0.75) circle (0.05) node (samb) {};
  \draw (-1.24,0.75) circle (0.2);
  
  \draw (-0.88,0) rectangle (-1.12,0.24) node [yshift=-0.16cm] (oamb) {};
  \draw (-1.12,0.24) -- (-1.75,0.24);
  \draw (-1.12,0) -- (-1.5,0);

  \draw [context, dotted] (-0.68,-0.2) rectangle node [sonode] {Ambulance}(-1.95,1.2);

  % inner communication
  \path [->] (rum) edge (tp)
  (tp.250) edge (nav.110)
  (nav.70) edge (tp.290)
  (ca) edge (tp)
  (ca.170) edge (mdb.10)
  (mdb.350) edge (ca.190);

  % outer communication
  \path [->,dashed]
  (seme) edge (ca)
  (ca) edge[solid] (oeme)
  (scd) edge (tp)
  (tp) edge[solid] (ocd)
  (sts) edge (nav)
  (samb) edge (rum.south)
  (tp.195) edge[solid] (oamb);
  % amcs border
  \draw (-0.5,-0.5) rectangle (5.5,5.5);
  \draw (3.5,5.25) rectangle node [font=\tiny] {CAET Management} (5.5,5.5);
\end{tikzpicture}
%%% Local Variables: 
%%% mode: latex
%%% TeX-master: "../paper"
%%% End: 
}
   \caption{The Computer-Aided Emergency Team Management \amcs}
   \label{fig:caetm}
 \end{figure}
\par Figure~\ref{fig:caetm} depicts the example \amcs which models such a \emph{Computer-Aided Emergency Team Management System} (CAET Management System).
Note that interaction with a human (\egc EM employee) is modelled as a pair containing an input stream and an output stream.
The system consists of the following contexts:
\begin{description}
\item[Case Analyser (ca)]
This context implements a computer-aided call handling system which assists an emergency response employee (ER~employee) during answering an emergency call.
The system utilises reasoning methods to choose which questions need to be asked based on previous answers.
In addition, it may check whether answers are inconsistent (\egc amniotic sac bursts when the gender is male).
For these purposes the case analyser context may also consult a medical ontology represented by another context.
The communication with the ER~employee is represented, on the one hand, as a sensor that reads the input of the employee and, on the other hand, by an output stream which prints the questions and results on a computer screen.

During the collection of all the important facts for this emergency case, the analyser computes the priority of the case and passes it to the task planner.
\item[Med Ontology (mo)]
This medical ontology can be realised, \egc by a description logic reasoner which handles requests from the case analyser and returns more specific knowledge about ongoing cases.
This information may be used for the prioritisation of the importance of a case.
\item[Task Planner (tp)]
This context keeps track of emergency cases.
Based on the priority and age of a case and the availability and position of ambulances it suggests an efficient plan of action for the ambulances to the (human) case dispatcher (cd).
%This dispatcher will be the chief of operations for the current shift.
The dispatcher may approve some of the suggestions or all of them.
If the dispatcher has no faith in the given plan of action, she can also alter it at will.
These decisions are reported back to the planning system such that it can react to the alterations and provide further suggestions.
Based on the final plan, the task planner informs the ambulance about their new mission.\\
The knowledge base of the context is an answer-set program for reasoning about a suggested plan.
It gets the availability and position of the ambulances by the ambulance manager.
In addition, the cases with their priority are provided by the case analyser.
With this information, the task planner gives the locations of the ambulances together with the target locations of the cases to a navigation system which provides the distances (\iec the estimated time of arrival (ETA)) of all the ambulances to all the locations.
\item[Amb Manager (am)]
The ambulance manager is a database, which keeps track of the status and location of ambulance units.
Each ambulance team reports its status (\egc to be on duty, waiting for new mission, \ldots) to the database (modelled by the sensor ``Ambulance'' (amb)).
Additionally, the car periodically sends GPS-coordinates to the database.
These updates will be pushed to the task planner.
\item[Navigation (na)]
This part of the \amcs gets traffic information (\egc congestions, roadblocks, construction zones, \ldots) to predict the travel time for each route as accurate as possible.
The task planner may push a query to the navigation system, which consists of a list of locations of ambulance units and a list of locations of target areas.
Based on all the given information this context will return a ranking for each target area, representing the ETAs for each ambulance.
\end{description}
Now we want to have a closer look on the instantiation details of some aspects of our example.
At first we investigate the \cc relation of the case analyser.
It allows for the computation of new belief sets whenever the ER~employee pushes new information to the analyser.
In addition, it will also approve of a new computation if the medical ontology supplies some requested information.
Recall that the case analyser also assigns a priority to each case
and that we want to allow the employee to set the priority manually.
Let us suppose that such a manual override occurs and that the case analyser has an ongoing query to the medical ontology.
Due to the manual priority assignment, the requested information from the ontology is no longer needed.
Therefore, it would be desirable that \cc does not allow for a recomputation if all conclusions of the ontology are only related to the manually prioritised case.
With the same argumentation in mind, the context update function \cuf will also ignore this information on the input stream.
This kind of behaviour may need knowledge about past queries which can be provided by an additional output rule for the case analyser which feeds the relevant information back to the context.

Next, we will have a look at the task planner that is based on answer-set programming.
We will only present parts of the program, to show how the mechanics are intended to work.
To represent the incoming information on the input stream, the following predicates can be used:
\begin{description}
  \ttitem{case(caseid,loc,priority)}{represents an active case (with its location and priority) which needs to be assigned to an ambulance.}
  \ttitem{avail(amb,loc)}{states the location of an available ambulance.}
  \ttitem{eta(caseid,amb,value)}{provides the estimated time of arrival for a unit at the location of the target area of the case.}
  \ttitem{assign(amb,caseid)}{represents the assignment of an ambulance to a case by the dispatcher.}
\end{description}
These predicates will be added by the context update function to the knowledge base if corresponding information is put on the input stream of the context.
Based on this knowledge, the other components of the answer-set program will compute the belief sets (\egc via the stable model semantics).
Note that an already assigned ambulance or case will not be handled as an available ambulance or an active case, respectively.
In addition, \cuf can (and should) also manage forgetting of no longer needed knowledge.
For our scenario it may be suitable to remove all \texttt{eta}, \texttt{avail} and \texttt{case}  predicates when the cases or the unit is assigned.
The \texttt{assign} predicate can be removed when the ambulance manager reports that the assigned ambulance is available again.

The set \outr of output rules of the task planner could contain
the following rules:\footnote{Keep in mind that in an actual implementation one may want to provide further information via communication.}
\begin{eqnarray*}
  \tuple{\text{cd,assign}(A,C)} \la& \text{sugassignment}(A,C)\\
  \tuple{\text{na,queryA}(L)} \la& \text{avail}(A), \text{not assign}(A,\_), \text{loc}(A,L)\\
  \tuple{\text{na,queryC}(L)} \la& \text{case}(C,P), \text{loc}(A,L), \text{not assign}(A,\_)\\
  \tuple{\text{amb,assigned}(A,C)} \la& \text{assign}(A,C)
\end{eqnarray*}
The first rule informs the case dispatcher (cd) about a suggested assignment that has been computed by the answer-set program.
Rules two and three prepare lists of ambulances and cases for querying the navigation context.
Recall that the latter needs a list of ambulance locations (generated by rule two) and a list of target area locations (generated by rule three).
Also keep in mind that for each belief set a data package with all information for one context or output stream is constructed.
So the whole list of current target areas and free ambulance units will be passed to the navigation context at once.
The last rule notifies the ambulance team that it has been assigned to a specific case.

Related to this example we want to mention privacy aspects as a real world policy which is especially important to applications in public services and health care.
As the multi-context system is a heterogeneous system with different contexts, a completely free exchange of data may be against privacy policies.
This issue can be addressed by the adequate design of output rules, which can also be altered with respect to additional information in the input stream (\egc some context gains the permission to receive real names instead of anonymous data).
So each context may decide by its own which parts of the belief sets are shared and exchanged with other contexts.

Another interesting aspect about \amcss is the possibility to easily join two \amcss together, outsource a subset of contexts in a new \amcs, or to view an \amcs as an abstract context for another \amcs in a modular way.
This can be achieved due to the abstract communication by means of streams.
With respect to our scenario there could be some \amcs which does the management of resources for hospitals (\egc free beds with their capabilities).
The task planner might communicate with this system to take the needed services for a case into account (\egc intensive care unit) and informs the hospital via these streams about incoming patients.
It would be easy to join both \amcss together to one big system or to outsource some contexts as input sensors paired with an output stream.
In addition, one may also combine different contexts or a whole \amcs to one abstract context to provide a dynamic granularity of information about the system and to group different reasoning tasks together.

\section{Relation to Reactive Multi-Context Systems}\label{sec:mmcs}
In this section we want to address differences and communalities between \amcss and \rmcss~\cite{Brewka2014} 
as both are types of multi-context systems that work in an online fashion and can react to external information.
Runs of \rmcss are based on equilibria which are collections of belief sets---one for each context---on which,
intuitively, all of the contexts have to agree.
Thus, equilibria can be seen as a tight integration approach in which the semantics of the individual contexts
are interdependent.
However, the high level of integration also comes at the price that the different contexts must wait for each other 
for the computation of each equilibrium, \iec they are synchronised.
In \amcss, on the other hand, the coupling of the semantics is much looser---communication between contexts only
works via data packages that are sent to another context after a computation
and not via a higher-level common semantics for multiple contexts.
But as a benefit, each context can run at its own pace which is useful in settings where there is a context
that requires much more time for evaluating its semantics than others.

A further difference is the role of non-determinism in the semantics of \amcss and \rmcss.
An equilibrium in an \amcs consists of a single belief set for each context.
Hence, as \amcss also use a multiple belief set semantics,
there may also be multiple equilibria as a source of non-determinism at each step in a run.
For \amcss, all belief sets of a context are computed in a consecutive way (we assume that if only a single belief set is desired than the semantics of the respective context should be adapted accordingly by the knowledge engineer).
Nevertheless, there is also a source of non-determinism in the case of \amcss caused by the undefined
duration of computations.

Regarding the computational complexity of the two frameworks,
the computation of an equilibrium requires
to guess an equilibrium candidate first before the semantics of the context is computed which 
is expensive regarding runtime when put to practice. 
In theory, this guess does not add extra complexity if the context semantics is already $\mathbf{NP}$-hard (as shown in~\cite{Brewka2014}) because it can be combined with the guesses required in the contexts. However, this trick cannot be used in implementations that uses black boxes for computing context semantics.
On the other hand, \amcss do not add substantial computational requirements to the effort needed for computing context semantics.
In particular, \amcss are scalable as adding a further context 
has no direct influence on how the semantics of the other contexts are computed but can only influence the input they get.

Both, \amcss and \rmcss are very general frameworks that 
allow for simulating Turing machines and thus for performing multi-purpose computations
even if only very simple context formalisms are used (if the length of a run is not restricted).
In this sense the approaches are equally expressive.
Moreover, when allowing for arbitrary contexts one could trivially simulate the other by
including it as a context.
Despite the existence of these straightforward translations, we next sketch how we simulate an \rmcs with an \amcs using a more direct translation, as this gives further insight into the differences
of the two frameworks. Moreover, it demonstrates a way to implement \rmcss by means of \amcss.
For every context $C_i$ of a given \rmcs $M_r$, we introduce three contexts in the \amcs~$M_a$ that simulates $M_r$:
\bi
 \item a context $C_i^{kb}$ that stores the current knowledge base of the context,
 \item a context $C_i^{kb'}$ in which a candidate for an updated knowledge base can be written and its semantics can be computed, and
 \item a management context $C_i^{m}$ that implements the bridge rules, and the management function of the context.
\ei
There are three further contexts: 
\bi 
\item $C^{obs}$ receives sensor data and distributes it to every context $C_i^{m}$ where $C_i$ depends on the respective sensor.
The context is also responsible for synchronisation: for each sensor, new sensor data is only passed on after an equilibrium has been computed.
\item $C^{guess}$ guesses equilibrium candidates for $M$ and passes them to the management contexts $C_i^{m}$. Based on that
and the information from $C^{obs}$, $C_i^{m}$ computes an update $kb_i'$ of the knowledge base in $C_i^{kb}$ and stores $kb_i'$ in 
$C_i^{kb'}$. The latter context then computes the semantics of $kb_i'$ and passes it to the final context
\item $C^{check}$ that compares every belief set it receives with the equilibrium candidate (that it also receives from $C^{guess}$).
      If a matching belief set has been found for each context of $M_r$, the candidate is an actual equilibrium.
      In this case $C^{check}$ sends the equilibrium to an output stream and notifies the other contexts about the success.
\ei      
In case of a success, every context $C_i^{m}$ replaces the knowledge base in $C_i^{kb}$ by $kb_i$ and a next iteration begins.
In case no equilibrium was found but one of the $C_i^{kb'}$ contexts has finished its computation,
$C^{check}$ orders $C^{guess}$ to guess another equilibrium candidate.

\section{Related Work and Discussion}\label{sec:conclusion}
A concept similar to output-rules has been presented in the form of reactive bridge rules~\cite{Ellmauthaler2013a}.
There the flow of information is represented by rules which add knowledge to the input streams of other contexts.
Which information is communicated to other contexts is also determined by the local belief set of each context.

Note that evolving multi-context systems~\cite{GKL2014} follow a quite similar approach as \rmcss and hence
the relation of \amcss to \rmcss sketched in the previous section also applies in essence to this approach.

The system \texttt{clingo}~\cite{Gebser2012} is a reactive answer-set programming solver.
It utilises TCP/IP ports for incoming input streams and does also report the resulting answer sets via such a port.
It provides means to compute different semantics and can keep learned structures and knowledge from previous solving steps.
Although there are no output rules or input stream pre-processing as in \amcss, the system features embedded imperative programming languages which may be helpful to model some of the presented concepts of this paper. 

% The system \texttt{clingo}~\cite{Gebser2012} is a reactive answer-set programming solver which implements some of the presented ideas.
% This tool has the capability to deal with an input stream where the incoming information may be preprocessed.
% Additionally, it provides means to compute different semantics and keep learned structures and knowledge from previous solving steps.
% As the solver can handle the input stream by its own it can be viewed as a realisation of parts of the context manager \cm.
% So some functionality of \texttt{clingo} is an instance of the context update function and the response of answer sets 
% %(which may be projected)
%  can be seen as a form of output rule handling.

In general, the tasks performed by a context management can be realised by different formalisms (\egc imperative scripting languages or declarative programming). Here, it seems likely that different languages can be the most appropriate management language, depending on the type of context formalism and the concrete problem domain.
A feature that is not modelled in our proposal but that is potentially useful and we intend to consider in the future is to allow for aborting computations.
Moreover, we want to study modelling patterns and best practices for \amcss design for typical application settings and compare different inter-context topologies and communication strategies. %, \egc a context querying another context and waiting for the results versus a context that only triggers another context

The next natural step towards an implementation is an analysis of how existing tools such as \texttt{clingo} could be used for a realisation.
It is clear that such formalisms can be used as a context formalism.
Moreover, we are interested in how reactive features of \texttt{clingo} (\egc iterative computation, on-demand grounding, online-queries, \ldots) relate to \amcs concepts (\egc \cc, \ib, \ldots)
and whether the system can be described in terms of an \amcs.

% The next natural step towards an implementation is an analysis of how existing tools such as \texttt{clingo} could be used for a realisation.
% Clearly, such formalisms can be used as a context formalism.
% Moreover, we are interested in how these features for reactivity (\egc iterative computation, on-demand grounding, online-queries, \ldots) relate to parts of other \amcs concepts (\egc \cc, \ib, \ldots).
% This will also lead to a of specification of such an formalism with an \amcs.
%In a further step it may be useful to use system descriptions in terms of an \amcs to provide a collection of distinct and uniform specifications about the capabilities of existing systems.
%So it get easier to identify which parts of a task shall be handled by a given context formalism and which need to be captured and realised by the \amcs concepts.

% In a further step it may be useful to use \amcss and specifications to provide a framework for specifications and development of modular usable contexts.
% The basic idea is to provide a framework which clearly distinct which parts of a task shall be handled by the context formalism and which need to be realised by the \amcs environment. This may help to reuse contexts, exploiting the modular properties of \amcss further, and make their use for non-researchers more convenient.

% link to read: http://www.dblp.org/search/index.php#query=barbieri 2010 sparq

\end{document}